\documentclass[conference]{ieeeconf}
\IEEEoverridecommandlockouts
\usepackage{amsmath,amsfonts,amssymb}
\usepackage{algorithm}
\usepackage{array}
\usepackage[caption=false,font=normalsize,labelfont=sf,textfont=sf]{subfig}
\usepackage[font=footnotesize,labelfont=bf]{caption}

\usepackage{textcomp}
\usepackage{stfloats}
\usepackage{url}
\usepackage{verbatim}
\usepackage{graphicx}
\usepackage{cite}
\usepackage{booktabs}
\usepackage{colortbl}
\usepackage{pifont}

\usepackage{algpseudocode}
\usepackage{adjustbox}
\usepackage{tabularx}
\usepackage{makecell}
\usepackage{xcolor} 
\usepackage[colorlinks=true,
            linkcolor=blue,
            citecolor=blue,
            urlcolor=blue]{hyperref}
\providecommand{\methodname}{\texttt{SymSkill}}

\providecommand{\pose}{\ensuremath \mathbf{T}}
\providecommand{\type}{\ensuremath \mathbf{\lambda}}

\providecommand{\alltypes}{\ensuremath \Lambda}

\providecommand{\allobj}{\ensuremath \mathcal{O}}
\providecommand{\allframes}{\ensuremath \mathcal{F}}
\providecommand{\state}{\ensuremath  \mathbf{x}}

\providecommand{\pred}{\psi}

\providecommand{\dspolicy}{\ensuremath f}
\providecommand{\dsparams}{\ensuremath \mathbf{\Theta}}
\providecommand{\operator}{\ensuremath \alpha}

\providecommand{\objref}{ {o_\textrm{ref}}}
\providecommand{\objint}{ {o_\textrm{int}}}

\providecommand{\datamotion}{\mathcal{D}_\textrm{motion}}
\providecommand{\datapre}{\mathcal{D}_\textrm{pre}}

\newcommand{\stkout}[1]{\ifmmode\text{\sout{\ensuremath{#1}}}\else\sout{#1}\fi}

\begin{document}

\title{SymSkill: Symbol and Skill Co-Invention for Data-Efficient \\ and Reactive Long-Horizon Manipulation}

\author{
Yifei Simon Shao, Yuchen Zheng, Sunan Sun, Pratik Chaudhari, Vijay Kumar and Nadia Figueroa \\
GRASP Laboratory, University of Pennsylvania, Philadelphia, PA, 19104 USA\\
{yishao, zhengyc, sunan, pratikac, kumar, nadiafig}@seas.upenn.edu
}

\markboth{2025 CORL Learning Effective Abstractions for Planning Workshop}%
{SymSkill: Symbol and Skill Co-Invention for Data-Efficient and Real-Time Long-Horizon Manipulation}


\maketitle

\begin{abstract}
Multi-step manipulation in dynamic environments remains challenging. Imitation learning (IL) is reactive but lacks \emph{compositional generalization}, since monolithic policies do not decide which skill to reuse when scenes change. Classical task-and-motion planning (TAMP) offers compositionality, but its high \emph{planning latency} prevents real-time failure recovery. We introduce \texttt{SymSkill}, a unified framework that jointly learns predicates, operators, and skills from unlabeled, unsegmented demonstrations, combining compositional generalization with real-time recovery. Offline, \texttt{SymSkill} learns symbolic abstractions and goal-oriented skills directly from demonstrations. Online, given a conjunction of learned predicates, it uses a symbolic planner to compose and reorder skills to achieve symbolic goals while recovering from failures at both the motion and symbolic levels in real time. Coupled with a compliant controller, \texttt{SymSkill} supports safe execution under human and environmental disturbances. In RoboCasa simulation, \texttt{SymSkill} executes 12 single-step tasks with 85\% success and composes them into multi-step plans without additional data. On a real Franka robot, it learns from 5 minutes of play data and performs 12-step tasks from goal specifications. Code and additional analysis are available at \url{https://symskill.github.io/}.
\end{abstract}

\section{Introduction}
Enabling robots to perform complex, long‑horizon manipulation in the real world remains challenging. Recent imitation‑learning (IL) approaches \cite{chi2023diffusion, zhao2023learning} excel at reproducing skills given large, high‑quality datasets, but tend to learn monolithic policies rather than reusable skills and predicates that compose into multi‑step plans. Historically, Task and Motion Planning (TAMP) bridges this gap by decomposing problems into symbolic planning over predicates/operators and continuous motion generation \cite{garrett2021integrated}. However, two factors limit TAMP scalability in practice. 1) Symbols and skills are often hand‑engineered and tuned per environment, which is labor‑intensive. 2) TAMP takes tens to hundreds of seconds to solve a large problem in a realistic contact-rich simulation environments\cite{shah2024reals}, making it infeasible to plan in dynamic environments with moving objects, or achieve real-time failure recovery at the symbolic or motion level.

\begin{figure}[!ht]
  \centering
  \includegraphics[width=\linewidth]{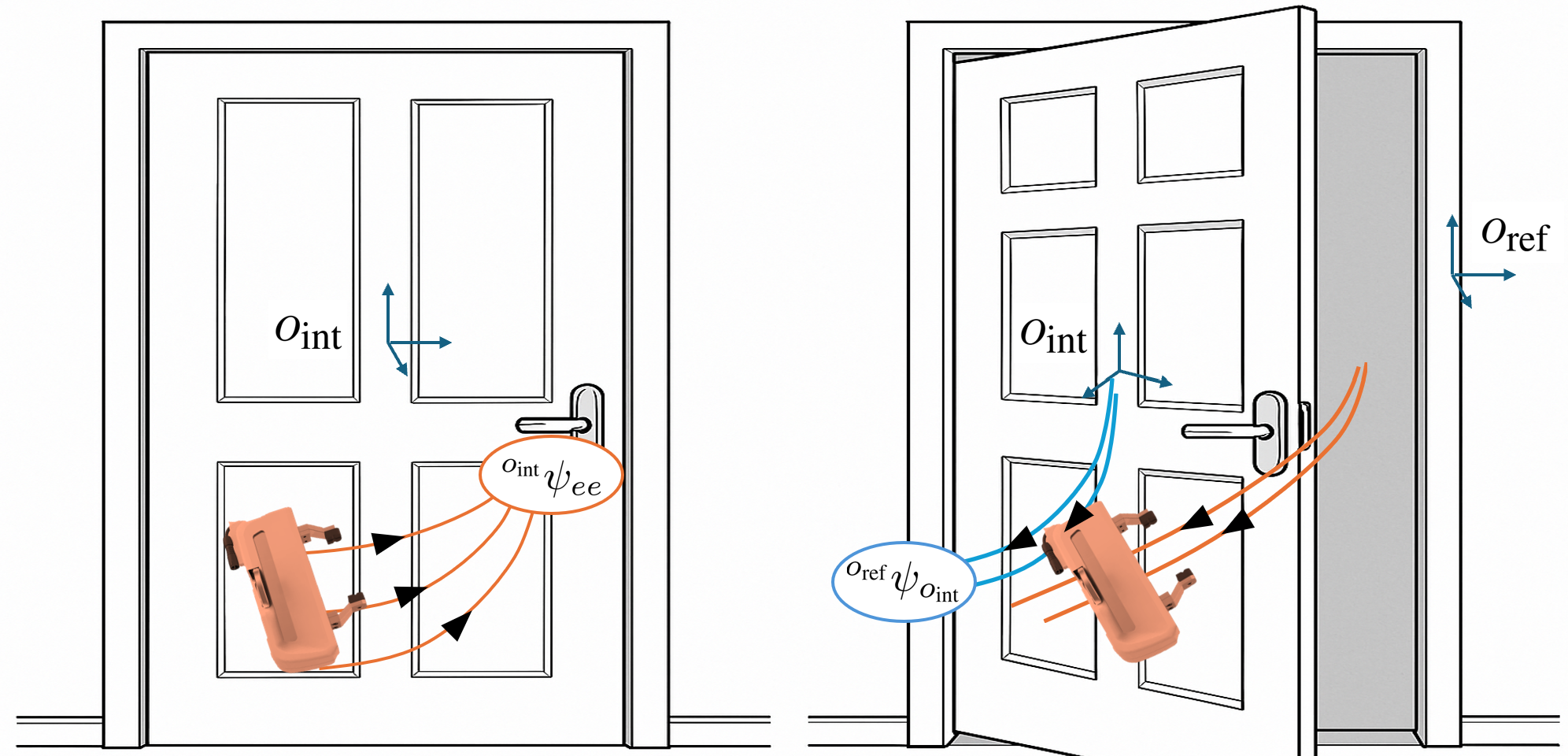}
  \caption{\small{
  Illustration of the \methodname{} predicate and skill co-invention process on a DoorOpen task. 
  \textbf{Left:} In the \texttt{premotion} segment (end-effector only motion), the object in motion in the next segment is treated as the object of interest $\objint$, and its frame serves as the reference for both predicate and skill learning. End-effector trajectories in this frame are used to fit SE(3) LPV-DS skills, and their endpoints are clustered to yield object-gripper relative pose predicates ${}^{\objint}\psi_{ee}$.
  \textbf{Right:} In the \texttt{motion} segment (gripper + object moving), a reference object $\objref$ is selected by querying a VLM on frames from the segment. Gripper trajectories are then expressed in the $\objref$ frame and used to fit a DS skill. Endpoints of the manipulated object trajectory in the $\objref$ frame are clustered to yield object–object relative pose predicates ${}^\objref\psi_\objint$. 
  }\label{fig:predicate_skill_coinvention}}
  \vspace{-15pt}
\end{figure}

\textit{Symbol and Skill Co-Invention} methods, such as \cite{keller2025neuro}, aim to combine the benefits of IL and TAMP by learning reusable symbols and skills from robot demonstrations and planning over them at runtime. As shown in \cite{silver2023predicate, keller2025neuro}, there is a delicate trade-off between inventing long-horizon operators that are too general to support effective planning and inventing operators that are too fine-grained to admit robust skill learning from limited data. Existing methods therefore rely on propose-and-down-select optimization procedures for predicate selection, but these searches can become slow as the number of objects and demonstrations grows, and may still fail to recover semantically meaningful predicates. We instead exploit a simpler structural prior: many manipulation interactions follow a small number of recurring object-centric patterns, where robots approach an object through a limited set of relative poses and manipulated objects come to rest in a small number of meaningful poses relative to nearby reference objects. Inspired by recent works that use VLMs to identify task-relevant objects, we use a VLM only in a lightweight offline role to identify the relevant stationary reference object in each demonstration, enabling relative-frame predicate and skill learning without relying on VLMs for online reasoning or control.


\begin{figure*}[htbp]
    \centering
\includegraphics[width=\linewidth]{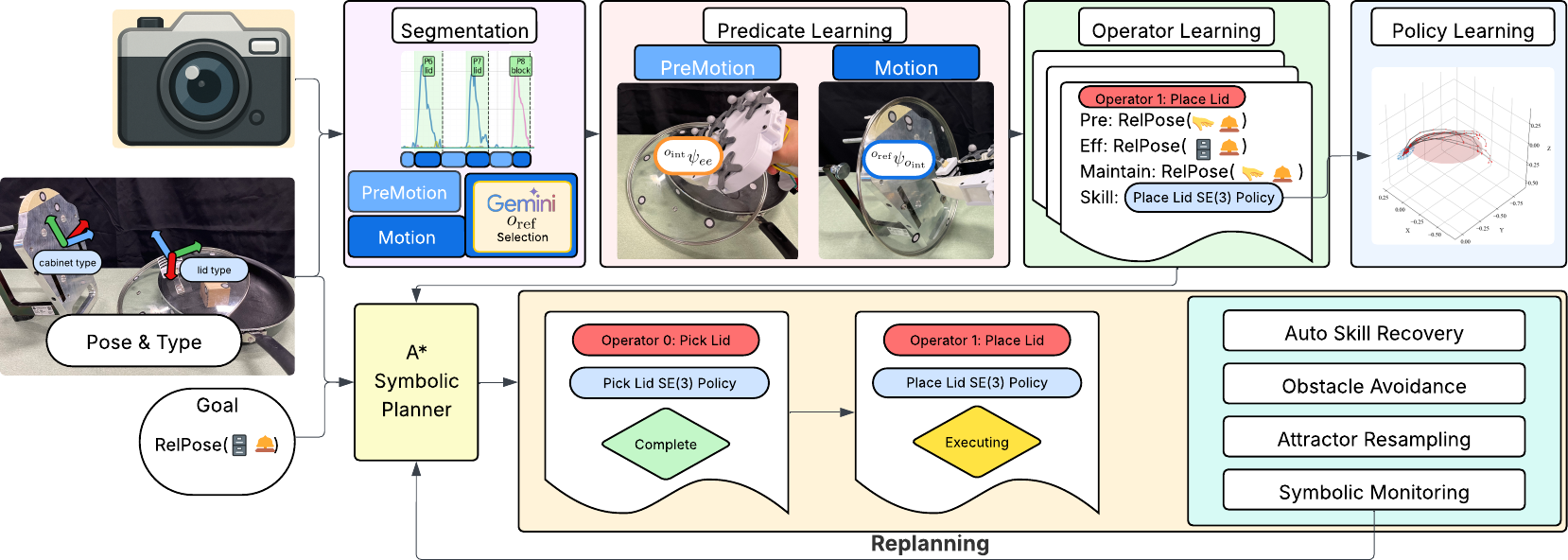}
\caption{\methodname{} offline pipeline (top half) and the online pipeline (bottom half). Subsection \ref{subsec: demo_seg_ref} (purple) describes segmentation and reference frame selection. Subsection \ref{sec:symbol-learning} (orange) describes how predicates are learned for each segment. Subsection \ref{sec: operator_learning} (green) learns the operators for online planning. Subsection \ref{sec: skill_learning} (blue) describes how each operator's skill is learned. Subsection \ref{sec: online} (yellow) describes how \methodname{} operates online. \label{fig: offline_online_pipeline}}
\vspace{-18pt}
\end{figure*}

To this end, we propose \texttt{SymSkill}, a unified framework that learns \textit{predicates}, \textit{operators}, and \textit{goal-oriented skills} in an unsupervised manner from unsegmented robot demonstration data, requiring as few as 5 demonstrations per task. At the symbolic level, \methodname{} identifies the object each trajectory segment moves toward using a VLM and automatically defines predicates as relative-pose classifiers. At the motion level, we adopt a dynamical-system (DS)-based approach to learn stable motion policies from minimal demonstration data.  
At execution time, given a symbolic goal specified by the learned predicates, \methodname{} uses a symbolic planner to compose skills into long-horizon plans that generalize across goals. Because replanning occurs only at the symbolic level, \methodname{} supports real-time error recovery. 
Together with the learned DS skills and a compliant passive DS controller, \methodname{} is robust to continuous-state perturbations without requiring replanning.
In RoboCasa, \methodname{} achieves an 85\% success rate on single-step tasks. Without additional data, it composes these skills to perform multi-step tasks. We also validate the approach on a Franka Panda robot, where \methodname{} learns 11 operators from 5 minutes of play data and achieves user-specified symbolic goals in real time.

\textbf{Contributions} 1) a framework for joint discovery and learning of symbols and goal-oriented DS skills from unlabeled and unsegmented demonstrations of short and long-horizon tasks, 2) online execution and failure recovery with reactive planning at the task and motion level, and 3) an open-source implementation for out-of-the-box robot-learning in RoboCasa~\cite{nasiriany2024robocasa} with its original demonstrations.

\section{Problem Statement}
\label{sec:problem}
We consider the problem of learning from play in \textit{deterministic}, \textit{fully observed} manipulation domains.
Let $\allobj$ be the set of objects, where each object $o \in \allobj$ is assigned to a type $\type(o)$ drawn from a predefined finite set $\alltypes$.
Let $\allframes = \{ee\}\cup\{o:o\in\allobj\}$ denote the set of kinematic frames of end-effector and object frames.

A pose $\pose\in SE(3)$ comprises position and orientation; ${}^{A}\pose_{B}\in SE(3)$ denotes the pose of frame $B$ expressed in frame $A$.
At time $t$, the continuous state in world frame is
\[
\state_t \;=\; \Big(\pose_{ee},\, \{\pose_{(o)}\}_{o\in\allobj},\, \{\type(o)\}_{o\in\allobj}\Big).
\]
Consistent with related works that also assume access to complete object states in simulation or via fiducial-based perception systems \cite{keller2025neuro, silver2023predicate}, we assume an having a perception module that provides tracked 6D poses and object types for all task-relevant objects at each timestep. 

\textbf{Problem Setup.} We are given $N$ unlabeled and \emph{unsegmented} robot demonstrations,
\[
\mathcal{D}=\{\tau_i\}_{i=1}^N,\qquad
\tau_i=\big\{\state_t\big\}_{t=0}^{T_i}.
\]
Each $\tau_i$ of length $T_i$ contains one or more demonstration trajectories of arbitrary object manipulating in the scene. For each trajectory, we record a time-synchronized RGB video of the workspace that keeps all task-relevant objects in view. 
A low-level skill policy outputs a 6D end-effector twist,
\begin{equation}
\begin{bmatrix}v, \omega\end{bmatrix}^T= \dspolicy(\cdot),
\end{equation}
where $v, \omega$ are the linear and angular end-effector velocities, respectively. We further assume gripper action $g = \{\text{open}, \text{closed}\}$ to be either open or closed throughout the policy. 
At test time, given initial state $\state_0$ and goal state $\state_G$, we seek to apply sequentially a number of policy tuple $\langle\dspolicy, g\rangle$ so that the $\state_G$ is achieved, while monitoring and recovering from failure in real-time. The robot action defined by policy $f$ is tracked by the following passive impedance controller 
\begin{align}\label{eq:passiveDS}
    F_{ee} = G - D(\dot{\pose}_{ee} - f(\cdot)),
\end{align} 
where $G\in\mathbb{R}^6$ is the gravity compensation term, $\dot{\pose}_{ee}\in\mathbb{R}^6$ is the end-effector velocity and $D$ is the damping gain ensuring the control input is energy dissipating in the directions orthogonal to the desired velocities, as in \cite{kronander2015passive}.

\section{Preliminary}
\label{sec:prelim}
\subsection{Learning Stable SE(3) Policy in Relative Frame}
\label{sec: prelim-skill}
When learning skills, we use dynamical system-based motion policy\cite{Khansari-Zadeh_Billard_2012a, figueroa2018physically, TEXTBOOK}. By leveraging redundancy of solutions from demonstration data, a learned dynamical system (DS) can be used as a stable motion policy that is robust to both temporal and spatial uncertainty. Specifically, we implement SE(3) LPV-DS \cite{sun2024se3linearparametervarying} combined with convex policy learning\cite{li2025elastic}, which requires only a small amount of data and is used later to fit one stable skill policy per learned operator. The framework consists of a linear Parameter Varying DS (LPV-DS), $f_p$, for position control and a Quaternion-DS, $f_o$, for orientation control: 

\begin{equation}\label{eq:se3lpvds}
\begin{aligned}
        v = f_p(x;\ \Theta_p), \, \,\,\,\omega = f_o(\bold{q};\ \Theta_o),
\end{aligned}
\end{equation}
where the inputs are position $x \in \mathbb{R}^3$ and orientation $\bold{q} \in SO(3) $ represented as quaternions, and each function is parameterized by $\Theta_*$. 
Using the LPV-DS as an example, the function $f_p$ has the form of a mixture of continuous linear time-invariant (LTI) system:
\begin{equation} \label{eq:lpvds}
    v=\sum_{k=1}^K \gamma_k(x)\bold{A}_k \left(x-x^*\right),
\end{equation} where $K$ represents the total number of LTI systems and $\gamma_k(x)$ is the mixing function that assigns the weight of each LTI system. $\gamma_k(x)$ is characterized by the Gaussian Mixture Model (GMM) parameters $\{\pi_k, \mu_k, \bold{\Sigma}_k\}_{k=1}^K$, which are estimated by fitting a GMM to the reference trajectories. Subsequently, each LTI system $\bold{A}_k$ is learned by solving a semi-definite program (SDP) with constraints enforcing globally asymptotic stability. For more details on SE(3) LPV-DS, please refer to \cite{figueroa2018physically, TEXTBOOK, sun2024se3linearparametervarying}.

\subsection{Symbolic Abstraction and Task and Motion Planning}
A \emph{predicate}, in this work, is a function, $\psi(A, B)$, that takes a tuple of frames as input and maps to a truth value as, 
\begin{equation}
 \psi_{\lambda_1, \lambda_2}(A, B)\rightarrow \{\text{True},\text{False}\} 
\end{equation}
such that $\type(A) = \lambda_1$ and $\type(B) = \lambda_2$. Instantiating all predicates over all type-consistent tuples in $\state_t$ yields the symbolic state $s_t$ (the set of true ground atoms). 


An \emph{operator} $\operator = \langle \text{params}, \text{pre}, \text{eff}, \text{maintain}, \text{\emph{skill}} \rangle$ 
is a typed template defined over objects/frames. 
It consists of:  
(i) \textbf{parameters} $\text{params} = [\lambda_1, \lambda_2, \dots]$ specifying the required types of objects/frames,  
(ii) \textbf{preconditions} $\text{pre}(\operator)$, the set of predicates that must hold in the symbolic state before the operator can be executed,  
(iii) \textbf{effects} $\text{eff}(\operator)$, consisting of \emph{add effects} (predicates made true) and \emph{delete effects} (predicates made false) after execution, and  
(iv) \textbf{maintenance conditions} $\text{maintain}(\operator)$, the set of predicates that must hold throughout execution.  
(v) \textbf{\emph{skill}} a low-level policy tuple $\langle f, g\rangle$, such as the DS policy for $f(\cdot)$ (Sec.~\ref{sec: prelim-skill}) and grasping action $g$, that realizes this transition on the robot.  

Formally, a grounded operator defines a transition on symbolic states,
\begin{equation}
\operator([o_1,o_2,\dots], s_0) \rightarrow s_1,
\end{equation}
where each parameter is assigned a type-consistent object, i.e., $\lambda(o_i)=\lambda_i$. If the grounded preconditions are satisfied in $s_0$, the operator can be applied, producing $s_1$ through its effects while requiring its maintenance conditions to hold during execution.
The typical planning process is a slower than real-time search and optimization process, with methods like interleaved planning \cite{garrett2020pddlstream} or Search \& Sample (SeSame) \cite{mendez2023embodied}.
\begin{table*}[h]
\caption{Comparison of predicate and skill learning methods.} 
\label{tb: related_work}
\begin{adjustbox}{center}
\resizebox{\textwidth}{!}{%
\begin{tabular}{lllll}
\toprule
\textbf{Approach}                             & \textbf{Predicates}                                 & \textbf{Skills} & \textbf{\# of Demos} & \textbf{Planning Time}   \\ \hline
\rowcolor{blue!10} \textbf{SymSkill (Ours)} & Relative Pose Cluster (Start/End Motion) & SE(3) LPV-DS \cite{sun2024se3linearparametervarying}   & 1-10                       &  \textless{}100ms \\ 
NSIL \cite{keller2025neuro}                & Relative Pose Cluster (Low Relative Velocity)                  & MLP BC          & 200                      & \textless{}100ms \\ 
LAMP \cite{shah2024reals}                & Relational Critical Regions & Motion Planning (MP) & 200                      & \textgreater{} 50 s            \\ 
NOD-TAMP~\cite{cheng2024nod}                & NDF Features                  & Optimization + MP          & 1-10                      & \textgreater{} 50 s \\ 
\bottomrule
\end{tabular}%
}
\end{adjustbox}
\vspace{-18pt}
\end{table*}
\section{Related Work}
We categorize related work below, and compare the most relevant works to ours in Table \ref{tb: related_work}. Note that all methods in the table learn predicates in relative frames, which has proven a necessity for generalizable manipulation learning frameworks. Of all the methods, ours is the only one that plans in real time and requires fewer than 10 demonstrations.

\textbf{Data Generation for Visuomotor Policies:} Related to our approach are recent works that leverage relative frames for data generation \cite{xue2025demogen, mandlekar2023mimicgen, garrett2024skillmimicgen}. These methods typically segment human demonstrations into sub-trajectories and then \emph{stitch} them, either through simulation or direct perception editing, to augment data and train visuomotor policies from moderately sized datasets. While effective for scaling data, they do not learn the underlying \emph{task dependencies} from demonstrations, but instead reproduce rigid subtask sequences.



\textbf{Hierarchical Imitation Learning:} Current imitation learning (IL) strategies such as Diffusion Policy\cite{chi2023diffusion} excels at reproducing complex multi-modal skills, but they often degenerate on long-horizon tasks that require sequencing multiple skills. To address this, hierarchical IL methods \cite{wan2024lotus, wang2023mimicplay} decompose demonstrations into a high-level planner over skills and low-level controllers for execution. While this structure improves tractability and performance, the high-level planners provide no symbolic guarantees that the composed sequence of skills will achieve goal completion. Instead, their plans are statistical predictions from latent distributions, lacking logical verification or explicit reasoning over task dependencies.

\textbf{Symbol Learning with Skill Label:} One thread of work invents symbols with pre-defined skills and skill-labeled data \cite{kaelbling2017learning, konidaris2018skills, james2022autonomous, liu2024BLADE}. More recently, \cite{li2025bilevel} proposed to learn the neural effect predicates of operators and classifiers for these predicates together. \cite{cheng2024nod} (NOD-TAMP in Table \ref{tb: related_work}) uses NDF features \cite{simeonov2022neural} for learning grasping predicates. However, labeling is tedious, requiring teleoperating with pre-programmed skills, or performing direct operational-space teleoperation followed by skill labeling. 

\textbf{Symbol Learning with Unsegmented Data:} This class of methods propose a candidate pool of predicates using enumeration or VLM, and then sub-select using an objective function \cite{silver2023predicate, athalye2024predicate}.  When learning from a limited number of demonstrations or when the number of features for each object is high, these approaches often fail due to limited data or extended running time, as shown in Results. Notably, \cite{shah2024reals} (LAMP in Table \ref{tb: related_work}) proposes Relational Critical Regions (RCR) as predicates without performing the optimization. However, it still opts to use motion planning as the skill, making real-time failure recovery difficult. 

\textbf{Predicates/Operator/Skill Co-invention:} Among prior works, NSIL \cite{keller2025neuro} (NSIL in Table \ref{tb: related_work}) is the closest to ours. It uses relative low-velocity regions of the trajectory as meaningful candidate predicates. However, as shown in our experiment, this method fails to produce correct and semantically meaningful predicates and still requires the error-prone down-select optimization mentioned above.

\section{Methods}
\label{sec:method}
\methodname{} jointly learns predicates $\pred$, operators $\operator$, and skills $\dsparams$ from unsegmented demonstrations $\mathcal{D}$ and leverages them for real-time task execution. 
Demonstrations are segmented into end-effector–only \texttt{(premotion)} and end-effector–object \texttt{(motion)} segments, expressed in relative frames (Sec.\ref{subsec: demo_seg_ref}). From these segments, we cluster endpoints to invent relative-pose predicates (Sec.\ref{sec:symbol-learning}). Then the operators are derived by tracking predicate transitions (Sec.\ref{sec: operator_learning}). Lastly, DS policies skill for each operator is learned (Sec.\ref{sec: skill_learning}). At test time, symbolic goals are achieved by composing operators into skill sequences. Closed-loop DS policy ensures stability and disturbance rejection, while online monitoring and replanning enable real-time recovery. Fig.~\ref{fig: offline_online_pipeline} shows the offline and online pipeline of \methodname{}.

\subsection{Demo Segmentation and Reference-Frame Selection}
\label{subsec: demo_seg_ref}
We assume a demonstration $\tau=\{\state_t\}_{t=0}^{T}$ comprises of unordered episodes of skills, each with \texttt{premotion} $\rightarrow$ \texttt{motion} segments. 
A \texttt{premotion} segment is the motion of the end-effector gripper towards an object prior to making contact, while during a \texttt{motion} segment we assume at most one non-gripper object moves concurrently with the gripper. This holds in typical single-arm demonstrations for both rigid-object transport and single-joint articulated-object interactions. 

For each demonstration $\tau$, we compute linear and angular velocities for all frames $o$ and detect change points using a fixed threshold on either velocities.
For gripper end-effector $ee$ and object $o\in \mathcal {O}$, let $t^\text{start}$ and $t^\text{stop}$ denote the times at which some $o$ begins and ceases motion. We call this object the 
\emph{motion object} $\objint$ for that episode. We then extract two contiguous segments:
\[
\underbrace{\mathcal{S}^{\mathrm{pre}}_{\objint}=[t_0,\,t^\text{start})}_{\text{gripper-only motion (\texttt{premotion})}} 
\quad\text{and}\quad
\underbrace{\mathcal{S}^{\mathrm{mot}}_{\objint}=[t^\text{start},\,t^\text{stop}]}_{\text{gripper+object motion (\texttt{motion})}}.
\]
Here $t_0$ is the maximal time before $t^\text{start}$ such that no object other than $ee$ is moving in $[t_0,t^\text{start})$. 


For \texttt{premotion} segments, we express trajectories in the frame of the motion object and treat the frame of $\objint$ as the reference frame:
\[
\texttt{\texttt{premotion}:}\quad \bigl\{{}^{\objint}\pose_{ee}(t)\bigr\}_{t\in \mathcal{S}^{\mathrm{pre}}_{\objint}}.
\]

\begin{figure}[htbp]
\centering
\includegraphics[width=1\linewidth]{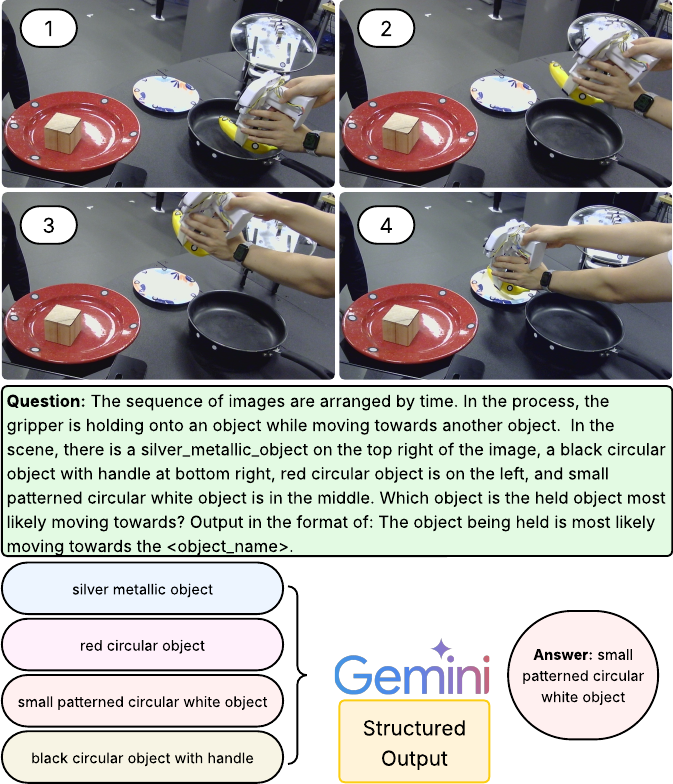}
\caption{The VLM prompt used for the real-world learning-from-play experiment proceeds as follows. First, the initial image is used to obtain text descriptions of all objects in view. Next, four equally spaced images from each \texttt{motion} segment are provided to Gemini together with the required output enumeration object, using the structured output feature. The returned text is then mapped back to the corresponding object name.
\label{fig:vlm prompt}}
\vspace{-18pt}
\end{figure}

For \texttt{motion} segments, both $ee$ and $\objint$ are in motion in the world frame. We do not assume rigid contact between them, since manipulating articulated items often involves non-prehensile movements. We assume $\objint$ motion is typically organized around one or a few \emph{reference objects}, each denoted as $\objref$ (e.g., transporting a cup into a sink, rotating a door w.r.t.\ its cabinet). To obtain all reference objects for \texttt{motion} segments individually for each motion episode, while capturing semantically meaningful reference objects, we query the Gemini-2.5-Pro\cite{comanici2025gemini} VLM on $n$ evenly spaced frames from $\mathcal{S}^{\mathrm{mot}}_{\objint}$ with a \emph{structured} output constrained to scene objects, as in Fig.~\ref{fig:vlm prompt}.
This structured output limits hallucination and enforces selection among known candidates.
With all $\objref$ fixed, we retain motion-segment trajectories in that frame:
\[
\texttt{motion:}\quad \bigl\{{}^{\objint}\pose_{ee}(t),\ {}^{\objref}\pose_{ee}(t),\ {}^{\objref}\pose_{\objint}(t)\bigr\}_{t\in \mathcal{S}^{\mathrm{mot}}_{\objint}}.
\]
We assume that objects of the same type can be manipulated in a similar manner, and that interactions between the same object type and reference type share common trajectory structures that can be exploited during learning. For now, we assume each object has a predefined type and $\type(\objint) \in \alltypes, \type(\objref) \in \alltypes$, but we can also easily expand to a open-object setting using a VLM for classification, as in \cite{athalye2024predicate}. 

\textbf{Outputs:} aggregating across demonstrations produces:
\begin{align}\label{eq:premotion_demo}
\datapre(\type_\objint) 
  &= \bigl\{\,\bigl({}^{\objint}\pose_{ee}(t)\bigr)_{t\in \mathcal{S}^{\mathrm{pre}}_{\objint}} 
     \;\big|\; \type(\objint) = \type_\objint \bigr\}
\end{align}
\begin{equation}
\label{eq:motion_demo}
\resizebox{0.91\hsize}{!}{%
$\begin{aligned}
 \datamotion\bigl(\type_\objint,\type_\objref \bigr)  &= \bigl\{\,\bigl({}^{\objint}\pose_{ee}(t),{}^{\objref}\pose_{ee}(t),  {}^{\objref}\pose_{\objint}(t)\bigr)_{t\in \mathcal{S}^{\mathrm{mot}}_{\objint}}\\
  &\big|\; \type(\objint) = \type_\objint 
           \cap \type(\objref) = \type_\objref \bigr\}.   
\end{aligned}$%
}
\end{equation}

\subsection{Relative Pose Predicate Learning}
\label{sec:symbol-learning}
We seek to capture distributions of relative poses that serve as meaningful symbolic predicates.
We consider the relative pose of the end-effector with respect to the motion object, ${}^{\objint}\pose_{ee}$, aggregated across $\datapre(\type_\objint)$. Rather than taking the last frame of each trajectory, which is unreliable under small datasets or non-prehensile motions, we fit normal distributions over the collection of poses observed in \texttt{motion} segments $\{{}^{\objint}\pose_{ee}(t)\}_{t\in \mathcal{S}^{\mathrm{mot}}_{\objint}}$. These are two independent Gaussians over translation ${}^{\objint}p_{ee}\!\sim\!\mathcal{N}(\mu^{\objint,ee}_{\text{pos}},\Sigma^{\objint,ee}_{\text{pos}})$ and orientation $\log({}^{\objint}R_{ee})\!\sim\!\mathcal{N}(\mu^{\objint,ee}_{\text{ori}},\Sigma^{\objint,ee}_{\text{ori}})$.
Given a new relative pose, we compute Mahalanobis distances to the respective means: $d_{\text{pos}}({}^{\objint}p_{ee}),\;\; d_{\text{ori}}(\log({}^{\objint}R_{ee}))$.
We declare the predicate ${}^{\objint}\psi_{ee}$ to hold if both distances $\epsilon_{\text{pos}},\epsilon_{\text{ori}}$:
\[
{}^{\objint}\psi_{ee}(\state) = \mathbf{1}\!\left[d_{\text{pos}} \le \epsilon_{\text{pos}} \wedge d_{\text{ori}} \le \epsilon_{\text{ori}}\right].
\]
fall below thresholds. Similarly, object–object relative pose predicates ${}^\objref\psi_\objint$ are obtained by fitting Gaussian distributions over $\{{}^{\objref}\pose_{\objint}(t)\}_{t\in \mathcal{S}^{\mathrm{mot}}_{\objint}}$, augmented with a short ($\approx$2s) post-motion window to stabilize end-pose estimation. The resulting ellipsoids not only define predicates but also serve as samplers for downstream goal-pose resampling during online recovery (Sec.~\ref{sec: online}).

\textbf{Outputs:}
Collecting these components yields the \emph{predicate libraries}
\[
\Psi_{\mathrm{pre}}(\type_\objint)=\{{}^\objint\psi_{ee}\}, 
\qquad
\Psi_{\mathrm{motion}}(\type_\objint,\type_\objref)=\{{}^\objref\psi_\objint\}.
\]

\subsection{Operator Learning using Learned Predicates}
\label{sec: operator_learning}
After we learn the relative‑pose predicates, we re‑evaluate all demonstration trajectories with $\Psi_{\text{pre}}(\type_{\objint})$ and $\Psi_{\text{motion}}(\type_{\objint},\type_{\objref})$ and invent symbolic operators using the method of \cite{chitnis2022learning}:
We first convert each demonstration into an abstract state sequence by evaluating all learned predicates at every \emph{demonstration segmentation boundary}. We denote the abstract states immediately before and after a transition as $s_0$ and $s_1$, respectively. 
Across these sequences, we identify recurring transition groups by finding segments with the same \emph{effects}, where effects are defined as
\[
\text{add}(\operator)=\bigcap_{(s_0,s_1)\in\mathcal{T}} (s_1 \setminus s_0), \qquad
\text{del}(\operator)=\bigcap_{(s_0,s_1)\in\mathcal{T}} (s_0 \setminus s_1),
\]
and
\[
\text{pre}(\operator)=\bigcap_{(s_0,s_1)\in\mathcal{T}} s_0.
\]
for each group of matched transition $\mathcal{T}$.
Because our system must monitor continuous states $\state$ online, we augment each operator with a set of \emph{maintain conditions} to be the intersection of all continuous‑state predicates that hold throughout the interval between $s_0$ and $s_1$,
\begin{align}
\text{maintain} \;=\; 
\bigcap_{\,t(s_0)\le t < t(s_1)} \!\!\state(t).
\end{align}
Together, we obtain a new operator
\[
\operator = \langle \text{params}, \text{pre}, \text{eff}, \text{maintain}, \text{\emph{skill}} \rangle ,
\]
where params($\operator$) are ordered and typed inputs that are automatically aggregated from all elements above.
 Tab.~\ref{tb:nsrt} shows the operators learned for the real-world learning-from-play experiment.
 
\textbf{Outputs:} We call the collection of operators $\Omega$, where each operator $\operator$ has trajectory segments from the dataset. Each operator's \emph{skill} will be learned in the next subsection. 

\begin{figure}[!tp]
    \centering
    \includegraphics[width=1\linewidth]{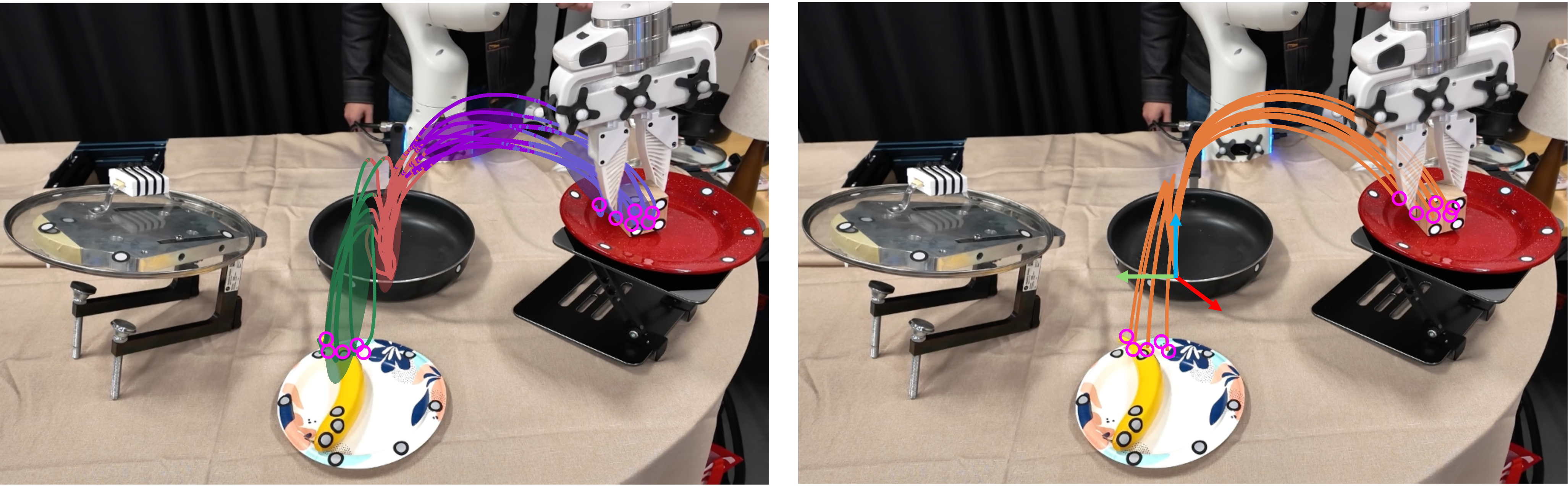}
    \caption{The visualization of demonstrations and SE(3) LPV-DS policy rollout for Op3 in Tab.\ref{tb:nsrt}. The left figure shows multiple collected trajectories placing a thing type item from various locations into the pan. The multimodal nature of the data is captured by 4 distinct Gaussians shown in different colors following the policy learning outlined in Sec.~\ref{sec: prelim-skill}. The right figure shows the reconstruction results of the learned policy starting from the same initial conditions, where the policy pose attractor in the pan frame is marked as an axis. All demonstrations converge on the attractor.
   \label{fig:policy_viz}}
    \vspace{-15pt}
\end{figure}

\subsection{SE(3) Skill Learning}
\label{sec: skill_learning}
Each operator $\operator \in \Omega$ requires a \emph{skill} = $\langle f,g\rangle$ for controlling the pose and gripper action of end-effector. We parameterize the policy $f_\alpha$ as a concatenated function of Eq.~\eqref{eq:se3lpvds}. 
For operators that model the \texttt{premotion} segments, we follow the learning procedures outlined in Sec.~\ref{sec: prelim-skill}, and use the demonstration data  $\{{}^{\objint}\pose_{ee}\}$ from Eq.~\eqref{eq:premotion_demo} to obtain the corresponding policy:
\begin{align}
    {}^\objint f_\operator({}^{\objint}\pose_{ee}; \Theta_p, \Theta_o).
\end{align}
For operators consisting of \texttt{motion} trajectories, the policies are expressed in $\objref$ frame following the same learning procedure using the demonstration data $\{{}^{\objref}\pose_{ee}\}$ from Eq.~\eqref{eq:motion_demo}:
\begin{align}\label{eq:policy_motion}
    {}^\objref f_\operator({}^{\objref}\pose_{ee}; \Theta_p, \Theta_o).
\end{align} For \texttt{motion} segments, specifically the ones including non-prehensile motions, we find that policies using relative pose trajectories between $ee, \objref$ perform significantly better than using relative pose trajectories between $\objint, \objref$, hence justifying the use of $\{{}^{\objref}\pose_{ee}\}$ in Eq.~\eqref{eq:policy_motion}. As introduced in Sec.~\ref{sec:problem}, the output of each learned policy is tracked by a task-space passive controller~\cite{kronander2015passive} as in Eq.~\eqref{eq:passiveDS}. One visualized policy is shown in Fig. \ref{fig:policy_viz}.

\subsection{Online Execution Monitoring and Adaptation} 
\label{sec: online}
The online algorithm requires a symbolic goal state $s_g$\footnote{$s_g$ is either directly specified or can be specified by symbolic abstraction of a goal state $\state_g$.}, expressed as a conjunction of one or more learned predicates. Given the current continuous state $\state_0$, we first compute its symbolic abstraction $s_0$. We then perform symbolic planning using A$^*$ search with the learned operators, producing a plan skeleton $\operator_1, \operator_2, \ldots, \operator_n$ from $s_0$ to $s_g$, if one exists. We then sequentially execute \emph{skill} in $\operator$, requiring little computation during execution. Since each \emph{skill} is a stable feedback policy, when $f_\operator$ outputs zero velocity, we advance to the next operator.

During execution, we monitor i) that the maintain conditions hold and ii) new expected effect satisfaction when each \emph{skill} ends. If failure occurs we replan from the current state. See project website for online algorithm. We summarize the elements that enable reliable recovery and eventual plan completion.

\textbf{Obstacle Avoidance} For each object in the scene $\allobj _{-\objint}$, excluding the ones that the gripper is holding or approaching, we model them as an ellipsoid and apply the local modulation introduced in \cite{Khansari-Zadeh_Billard_2012a} during skill execution: 
\begin{align}f' =\mathbf{M}(\mathcal{O}_{-\objint}) f(\pose; \dsparams),\end{align} where the modulated policy $f'$ incorporates the obstacle avoidance behavior and the modulation matrix $\mathbf{M}$ is constructed through eigenvalue decomposition with the normal and tangent directions of the defined ellipsoid boundaries.


\textbf{Resampling after failure} 
If a robot fails to execute a task on a given object, attempting it again without replanning will typically lead to another failure. Inspired by TAMP, a policy $f$ can be modified online by performing a frame transform when detecting failure during skill execution. Formally, we directly transform the policy: $f' = \pose f$, where $\pose$ is the pose sampled from the effect normal distribution as introduced in Sec.~\ref{sec:symbol-learning}.
1) When the maintain effect is lost, such as losing the grasp of an object, we assume the \textit{previous} skill needs to be resampled; 2) When effects of current $\operator$ is not satisfied at the end of \emph{skill}, we assume the attractor of the \textit{current} skill needs to be resampled. Therefore, depending on the operator sequence, we draw samples from ${}^\objint\psi_{ee}$ or ${}^\objref\psi_\objint$ to apply transformation. This strategy enables autonomous recovery from external disturbances, such as the robot regrasping a dropped object or reopening a closed cabinet door. 
\section{Experimental Results}
\label{sec:result}
We evaluate our method in \texttt{RoboCasa}\cite{nasiriany2024robocasa} simulation environment, and on the real Franka robot with motion capture and a webcam during learning.
\subsection{Single Step Simulation Result}
We exclusively use the demonstrations collected by the authors of the \texttt{RoboCasa} paper to ensure reproducibility. For single-step tasks, we reduce the variation in the demonstrations by filtering to keep only one variant of fixture per task, such as those \texttt{OpenSingleDoor} demonstrations with a cabinet that opens to the left. At test time, we also only generate environment with reduced task variation. Each task still have some randomness such as object initial poses. Table~\ref{tb:sim_result} shows the result of the proposed method by learning from 5-10 demonstrations per task: \textit{Proposed w/o monitoring} removes online maintain/effect checking and replanning, executing the policies in open loop once the initial symbolic plan is produced. \textit{Proposed w/ DP} shows \methodname{} when the low level policy is replaced by state-input U-Net-based Diffusion Policy (DP). 
\begin{table}[!tbp]
\centering
\caption{RoboCasa simulation result on 10 trials per task}
\begin{tabularx}{\columnwidth}{Xccc}
\toprule
\textbf{Task Success Rate \%} 
  & \textbf{Proposed} 
  & \makecell{\textbf{Proposed}\\\textbf{w/o Monitoring}} 
  & \makecell{\textbf{Proposed}\\\textbf{w/ DP}} \\ \toprule
\rowcolor{gray!7}OpenSingleDoor              & \textbf{100}  &  100 & 0 \\ 
CloseSingleDoor             &  \textbf{100} &  80 & 0 \\ 
\rowcolor{gray!7} PnPCounterToCab             &  \textbf{80} &   70& 0 \\ 
PnPCabToCounter             &  \textbf{100} &  40 & 0 \\ 
\rowcolor{gray!7} PnPStoveToCounter           &  \textbf{70} &   30& 0 \\ 
PnPCounterToStove           & \textbf{20}  &    0& 0 \\ 
\rowcolor{gray!7} OpenDrawer                  & \textbf{100}  &   100 & 0 \\ 
CloseDrawer                 & \textbf{70}  &    50& 40 \\ 
\rowcolor{gray!7} TurnOnStove                 & \textbf{100}  &   100& 0 \\ 
TurnOffStove                & \textbf{80}  &    30& 0 \\ 
\rowcolor{gray!7} TurnOnSinkFaucet            & \textbf{100}  &   100& 0 \\ 
TurnOffSinkFaucet           & \textbf{100}  &   90 & 0 \\ \bottomrule
\textbf{Average}             & \textbf{85.0}   &   \textbf{65.0}& {\textbf{3.3}} \\ \bottomrule
\end{tabularx}
\label{tb:sim_result}
\vspace{-18pt}
\end{table}

\begin{table*}[h!]
\centering
\caption{Learned Operators from play data: each couples symbolic transitions with SE(3) DS skills. Operators are arranged by semantic affinity.}
\resizebox{\textwidth}{!}{%
\begin{tabular}{lllll}
\toprule
\textbf{Operators} & \textbf{Human-Interpretable Summary} & \textbf{Preconditions} & \textbf{Effects} & \textbf{Maintain Conditions} \\ \toprule
\rowcolor{gray!7} Op7  & Pick lid \textit{from} cabinet & GripperOpen, Lid-in-cabinet & Gripper-in-lid, $\lnot$Lid-in-cabinet, $\lnot$GripperOpen & Lid-in-cabinet, GripperOpen \\ 
Op11 & Pick lid \textit{from} cookware & GripperOpen, Lid-in-cookware & Gripper-in-lid, $\lnot$Lid-in-cookware, $\lnot$GripperOpen & Lid-in-cookware, GripperOpen \\ 
\rowcolor{gray!7} Op1  & Place lid $\to$ cabinet & Gripper-in-lid & Lid-in-cabinet, $\lnot$Gripper-in-lid, GripperOpen & Gripper-in-lid \\
Op8  & Place lid $\to$ cookware & Gripper-in-lid & Lid-in-cookware, $\lnot$Gripper-in-lid, GripperOpen & Gripper-in-lid \\
\rowcolor{gray!7} Op9  & Pick thing \textit{from} drawer & GripperOpen, Thing-in-container, Thing-in-drawer & Gripper-in-thing, $\lnot$Thing-in-drawer, $\lnot$GripperOpen & Thing-in-container, Thing-in-drawer, GripperOpen \\
Op5  & Pick thing \textit{from} cookware & GripperOpen, Lid-in-cabinet, Thing-in-cookware & Gripper-in-thing, $\lnot$Thing-in-cookware, $\lnot$GripperOpen & Thing-in-cookware, Lid-in-cabinet, GripperOpen \\ 
\rowcolor{gray!7} Op10 & Pick thing \textit{from} container & GripperOpen, Thing-in-container & Gripper-in-thing, $\lnot$Thing-in-container, $\lnot$GripperOpen & Thing-in-container, GripperOpen \\
Op4  & Place thing $\to$ drawer & Gripper-in-thing, Thing-in-cookware & Thing-in-drawer, $\lnot$Gripper-in-thing, GripperOpen & Gripper-in-thing, Thing-in-cookware \\ 
\rowcolor{gray!7} Op3  & Place thing $\to$ cookware & Gripper-in-thing, Lid-in-cabinet & Thing-in-cookware, $\lnot$Gripper-in-thing, GripperOpen & Gripper-in-thing, Lid-in-cabinet \\ 
Op6  & Place thing $\to$ container & Gripper-in-thing & Thing-in-container, $\lnot$Gripper-in-thing, GripperOpen & Gripper-in-thing \\ \bottomrule
\end{tabular}}
\label{tb:nsrt}
\vspace{-10pt}
\end{table*}
\methodname{} correctly segments trajectories and identifies the object in motion. The VLM is almost always able to determine the reference object correctly. For \texttt{RoboCasa} tasks, we take the identified $\objint$ and $\objref$ and select the most frequent assignment across demonstrations, which yields perfect accuracy. Goal is specified by abstracting the symbolic state at the end of the majority of demonstrations.
Failure cases arise primarily in PnP tasks, where the randomly generated containers are sometimes too tall (e.g., a salad bowl). In such cases, the arm carrying the item collides with the container, causing task failure.

With only 5–10 demonstrations per task, DP is severely data-limited. In particular, \texttt{premotion} skills start from widely varying initial poses but occupy only a narrow funnel near the target object, so test-time states quickly fall out of distribution and execution often fails before reaching the object. Some low-variance \texttt{motion} skills can be reproduced qualitatively, but task success remains near zero because both phases must succeed. In contrast, the SE(3) LPV-DS skill defines a convergent feedback field in the learned reference frame, enabling steady progress toward the goal under perturbations.
We also evaluated DP with data augmentation from the DS policy, as detailed on the project website, but found no success with DP either. In contrast, the $\mathrm{SE}(3)$ LPV-DS controller induces a convergent vector field in the learned reference frame; its closed-loop stability prevents stalling and ensures steady progress to the goal even under perturbations.

For the symbol--skill co-invention baseline, we re-implemented NSIL~\cite{keller2025neuro} for qualitative comparison on \emph{OpenSingleDoor} and \emph{PnPCounterToCab}. In our low-data, multi-object setting, NSIL frequently selected spurious relative-pose predicates that explained the demonstrations but were not semantically meaningful for planning. The method was also sensitive to suboptimal demonstrations and slightly non-prehensile interactions, where useful contacts were often not included among the candidate predicates. As a result, NSIL failed to recover reusable predicates in these tasks, and the limited data was also insufficient for learning an effective low-level policy.

\subsection{Performing Multi-Step Task With No Additional Data}
We created a new task, \texttt{StoreCheese}, in RoboCasa. The task is successful when the robot picks the cheese from the cabinet, places it on the counter, and closes the cabinet door. To execute this task, we reuse only the previously learned predicates, operators, and skills from the constituent short-horizon tasks; we update operator preconditions via predicate evaluation across demonstrations. The operator from \emph{PnPCabToCounter} task thus has the predicate \emph{OpenSingleDoor-RelPose(Door, Cabinet)}, meaning door being open, as a precondition (illustrated as ${}^\objref\psi_\objint$ in Fig.\ref{fig:predicate_skill_coinvention}). We then manually specify the goal predicates as \{\emph{CloseSingleDoor-RelPose(Door, Cabinet)}, \emph{PnPCabToCounter-RelPose(Cheese, Counter)}\}. With this setup, the Franka robot successfully plans the operator sequence: open the door, pick and place the cheese, and finally close the door. It completes the task by chaining together six skills and recovering from symbolic errors multiple times. Video of the experiment can be found on the project website.

\begin{figure}[!tbp]
    \centering
    \includegraphics[width=1\linewidth]{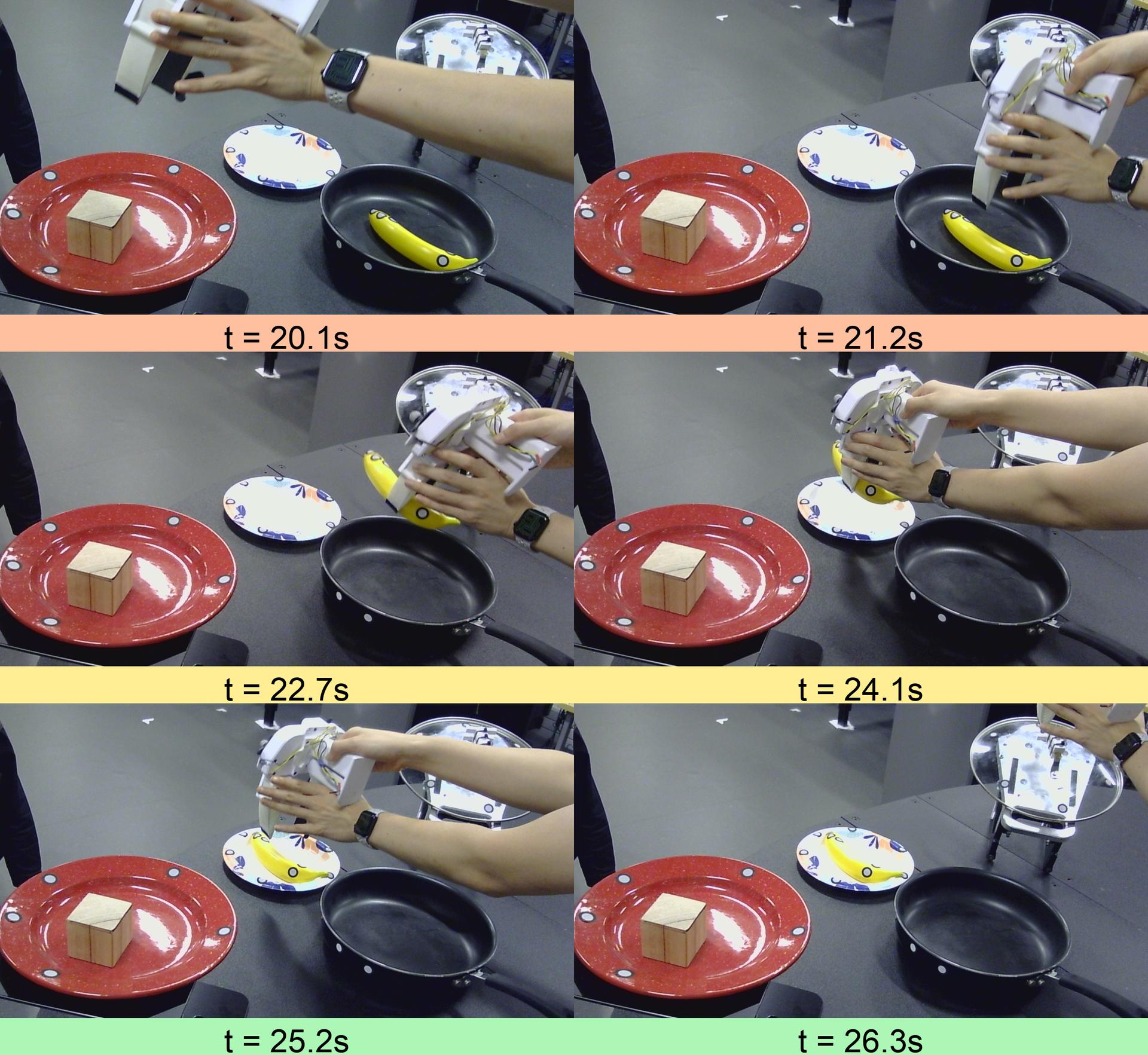}
    \caption{Real-world data collection pipeline. We use a motion capture system to record object interactions in the workspace. Here we show one motion episode with a sequence of timestamped images; the manipulated object ($\objint$) is a banana. Frames with orange, yellow, and green banners denote the \texttt{premotion}, \texttt{motion}, and post-motion segments, respectively.  \label{fig:real_world_demo}}
    \vspace{-20pt}
\end{figure}

\begin{figure}[!tbp]
    \centering
    \includegraphics[width=1\linewidth]{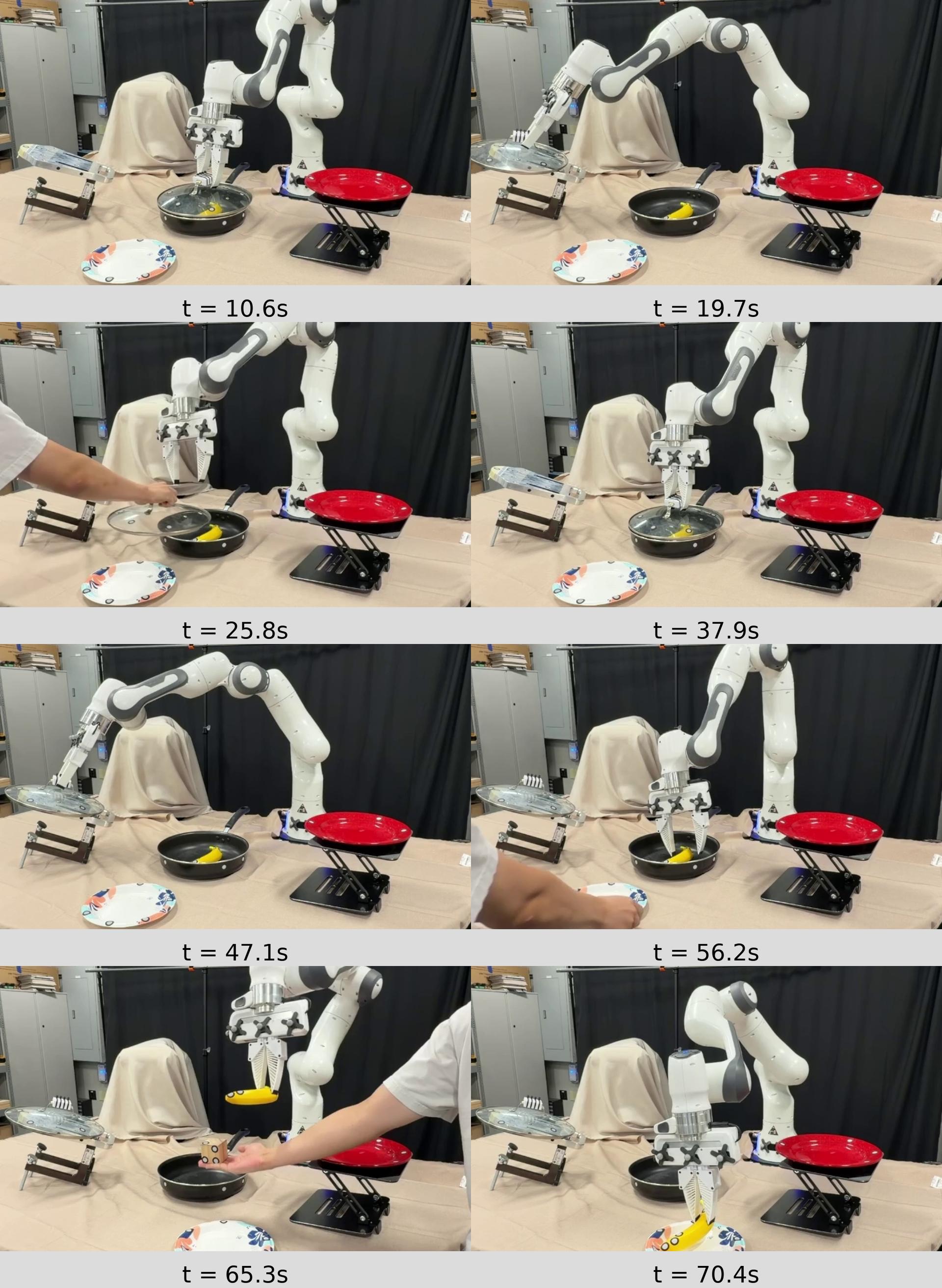}
    \caption{Real-world execution of \methodname{} toward the manually specified symbolic goal \texttt{\{RelPose(banana, plate)\}}. During execution, we introduce three disturbances to highlight the three recovery mechanisms of \methodname{}: closing the lid triggers symbolic-level recovery, moving the plate needs needs no explicit recovery with DS skills, and adding an obstacle is handled by modulation. \methodname{} successfully addresses all three and completes the task. \label{fig:real_world}}
    \vspace{-20pt}
\end{figure}

\subsection{Learning From Play In Real-World}
We demonstrate our method can learn from play data in the real world. We set up a scene with block and banana (\texttt{thing\_type}), red plate (\texttt{drawer\_type}), white plate (\texttt{container\_type}), dishrack (\texttt{cabinet\_type}), lid (\texttt{lid\_type}), and pan (\texttt{cookware\_type}). During play data collection, the demonstrator uses a UMI gripper\cite{chi2024universal} to perform sequences of manipulation tasks such as closing the pan with a lid or placing the banana on a plate. We obtain the pose of objects and the gripper from a motion capture system and record the video data from a webcam.
 Fig. \ref{fig:real_world_demo} shows the data collection process. Fig. \ref{fig:vlm prompt} shows selecting reference frame process and the VLM prompt. We find that with minimal prompt engineering, VLM can correctly identify the reference frame $\objref$, leading to correct learned predicates. Table \ref{tb:nsrt} summarizes the learned operators from approximately 5 minutes of unsegmented real-world play. 
Our method learns semantically meaningful and logical operators from unsegmented data, such as recognizing that picking items from the pan requires first removing the lid to place it on the dishrack. 
An example is shown in Fig.~\ref{fig:real_world}.

We also observe non-obvious but semantically consistent preconditions. For example, Op9 includes `Thing-in-container' as a precondition for picking from the red plate because this relation holds in all demonstrations. Although counterintuitive, this illustrates that \methodname{} captures dataset-specific task structure directly from play rather than relying on generic language priors.
We also demonstrate reacting to human external disturbance and recovering from failure in a \texttt{OpenSingleDoor} task. The video of the experiment is on the project website.


\section{Conclusion and Future Work}
\label{sec:conclusion}
We presented \methodname{}, a symbol–skill co-invention framework that jointly learns relative-pose predicates for planning and DS-based skills for execution. Our results in simulation and on real robots show that \methodname{} is significantly more sample-efficient and faster to learn than existing baselines, while enabling robust long-horizon manipulation. As future work, we plan to extend our framework to learn directly from egocentric video and to scale toward mobile manipulation scenarios, further broadening its applicability to real-world generalist robots.

\textbf{Acknowledgment}: We thank Bowen Li, Nishanth Kumar, Tom Silver and Rachel Holladay for the helpful discussions at various stages of the project. We thank Peng Qiu for helping out with setting up the simulator.

\appendices

\bibliographystyle{IEEEtran}
\bibliography{example}
\clearpage

\section{Online Algorithm}
\label{App: online_algo}
We show the pseudo-code of our online algorithm. 
\providecommand{\failmem}{failmem}
\providecommand{\abstr}{ \text{abstract}}
\providecommand{\goalpose}{ goal\_pose}
\begin{algorithm}[htbp] 
\footnotesize
\label{alg:online_execution}
\begin{algorithmic}[1] 
\Require Current State $\state_0$, Goal atoms $s_\text{g}$, Learned operators ${\Omega}$ (each operator $\operator$ = $\langle$ pre, maintain, eff, \emph{skill} $\rangle$), all objects $\allobj$
\State Initialize  replan\_count $\gets 0 $,$\failmem \gets \{\}$\Comment{failure memory}
    \State CurrentPlan$(\operator_1, \ldots, \operator_n)\gets\textproc{SymbolicPlanner}(s_{0}, s_{g}, \Omega)$ \label{line: online_replan}
    \If{$s_{g}$ unreachable or replan\_count $\ge 20$} 
    \State \Return Failure \EndIf
    \State $\operator_{exe} \gets \text{CurrentPlan}[i]$\label{line: online_loop}
    \State $\operator_{prev} \gets \text{CurrentPlan}[i-1]$
    \If {$\operator_{exe} \in \failmem$}
        \State Goal Pose $\gets$ sample(eff($\operator_{exe}$)) \Comment{Sample in ${}^{\objint}\psi_{ee}$ or ${}^{\objref}\psi_{\objint}$}
    \Else
    \State Goal Pose $\gets 0$
    \EndIf

        \While{maintain$(\operator_{exe})\in\abstr (\state)$}
        \State $<f,g> \gets$\emph{skill}$(\operator_{exe})$
        \State $f' \gets\mathbf{M}(\mathcal{O}_{-\objint}) f$ \Comment{Obstacle Avoidance}
        \If{$f'< \epsilon$} \Comment{Current skill ending}
            \If{eff($\operator_{exe}\notin\abstr (\state)$)}
                \State Add $(\operator_{exe})$ to $\failmem$ \Comment{Current skill failure}
                \State Go To Line: \ref{line: online_replan} \Comment{Current skill success}
            \EndIf
            \State $i \gets i+1$
            \State Go to Line: \ref{line: online_loop}
        \EndIf
        \EndWhile
    \State Add $(\operator_{prev})$ to $\failmem$ \Comment{Failure of maintain predicates}
    \State Go To Line: \ref{line: online_replan}
\end{algorithmic}
\end{algorithm}

\section{Extended Analysis of \cite{keller2025neuro}}
\label{app:nsil_analysis}
\paragraph{Difficulty 1: Assumption of Optimal Demonstrations.}
NSIL requires that candidate predicate sets yield plans whose length matches the demonstration length. In practice, demonstrations in RoboCasa often contain suboptimal behaviors, such as multiple approaches before grasping. For instance, in \texttt{OpenSingleDoor} or \texttt{PnPCounterToCab}, demonstrations sometimes include repeated approaches, leading to longer trajectories than the optimal plan. As a result, valid predicate sets are rejected because the demonstration is not optimal. We relaxed this by introducing a penalty for mismatched lengths rather than strict rejection, but the issue remains fundamental.

\paragraph{Difficulty 2: Ambiguity from Distractor Objects.}
Including irrelevant objects significantly increases ambiguity. For a simple pick-and-place task with one distractor, low-speed analysis generated 13 candidate predicates. Beam search often selected fragile ones, e.g., $\text{RelPose}(\text{DoorHandle}, \text{Object})$, because incorrect relationships (object to door) has the same cost as semantically meaningful ones (object to cabinet). Such predicates fail to generalize if the distractor moves. Our method mitigates this by using VLM-based semantic grounding to select reference objects, bypassing spurious associations.

\paragraph{Difficulty 3: Noisy Non-Prehensile Interactions.}
For articulated objects, grasping is frequently non-prehensile (e.g., pushing a door handle rather than firmly holding it). This makes $\text{RelPose}(\text{Gripper}, \text{Handle})$ a noisy predicate and artificially inflates the demonstration symbolic sequence length. In our trials with four demonstrations of door opening, the optimization incorrectly favored gripper-to-cabinet relations, which do not reflect the actual manipulation. This highlights NSIL’s difficulty in capturing non-prehensile skills robustly.

\paragraph{Summary.}
Overall, NSIL’s reliance on optimal demonstrations, its vulnerability to distractors, and its fragility in non-prehensile settings limit its robustness in realistic environments such as RoboCasa. By contrast, our framework leverages semantic grounding (via VLMs) for predicate discovery and SE(3) DS policies with stability guarantees for skill execution, making it more robust under limited demonstrations and realistic variations.

\section{DP with Data Augmentation}
\label{app:dp}
\begin{figure}[htbp]
\centering
\includegraphics[width=\linewidth]{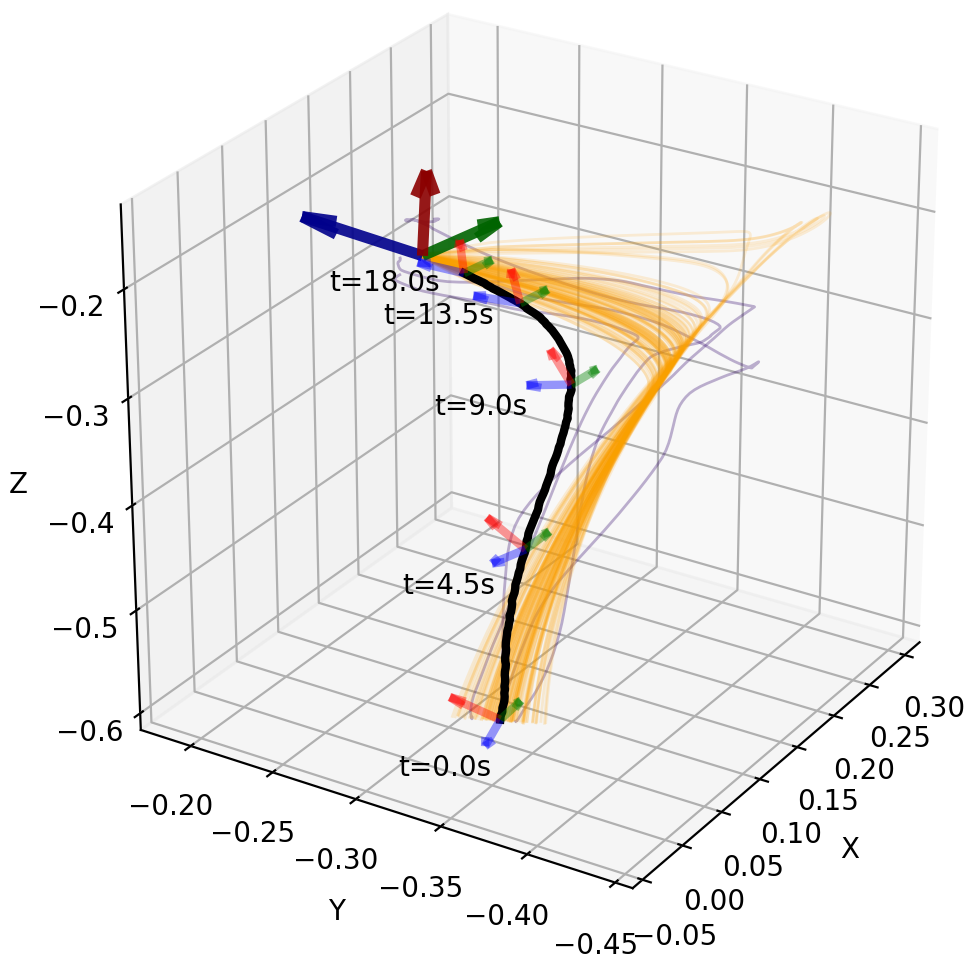}
\caption{DP rollout for the \texttt{premotion} segment of \textit{OpenSingleDoor}. Purple trajectories are 4 training demonstrations used for \methodname{}, yellow trajectories are augmented data generated by $\mathrm{SE}(3)$ LPV-DS, and the black curve is a rollout from the trained DP. The poses (red, green, blue axes) indicate end-effector orientation at several timestamps; the larger pose denotes the averaged final orientation across all demonstrations.}
\label{fig:dp-sim}
\end{figure}

We used the trained $\mathrm{SE}(3)$ LPV-DS policies to generate 100 rollouts for each skill as additional training data for the diffusion policy (DP). Initial end-effector poses were sampled from an $\mathrm{SE}(3)$ Gaussian fit to the initial poses in the training set. Example training data, augmented data, and a rollout generated by the trained DP are shown in Figure~\ref{fig:dp-sim}.

We used trained $\mathrm{SE}(3)$ LPV-DS policies to generate rollouts as additional training data for diffusion policy (DP). Initial end-effector poses were sampled from an $\mathrm{SE}(3)$ Gaussian fit to the initial poses in the training set. All training data, augmented data, and a rollout from the trained DP are shown in Figure~\ref{fig:dp-sim}.

We evaluated DP with data augmentation in \texttt{RoboCasa}. Although the rollout in Figure~\ref{fig:dp-sim} appears successful, we observed degraded orientation performance in simulation, leading to a zero success rate. We hypothesize this is due to the robot's kinematic constraints, which push the end-effector into regions outside the training distribution. DP succeeds on relatively simple tasks (e.g., \textit{CloseDrawer}) that do not require precise orientation control, but fails on tasks such as \textit{OpenSingleDoor} and \textit{PnPCounterToCab}, where the policy consistently approaches the $\objint$ but cannot achieve the grasp orientation required for success.

\end{document}